\documentclass[letterpaper, 10 pt, conference]{ieeeconf}  

\IEEEoverridecommandlockouts                              

\usepackage{graphicx} 
\usepackage{amsmath} 
\usepackage{amssymb}  
\usepackage{balance}
\usepackage{capt-of}
\usepackage{multirow}
\usepackage{booktabs}
\usepackage[font=small]{caption}
\usepackage{subcaption}
\usepackage{arydshln}
\usepackage{graphicx}
\usepackage{wrapfig}
\usepackage{url}

\usepackage{xcolor}
\definecolor{linkcolor}{RGB}{6,69,173}
\usepackage[
colorlinks=true,allcolors=linkcolor,pageanchor=true,plainpages=false,pdfpagelabels,bookmarks,bookmarksnumbered,backref=page
]{hyperref}

\usepackage[nameinlink,capitalise]{cleveref}
\crefname{figure}{Fig.}{Figs.}
\crefname{equation}{Eq.}{Eqs.}

\DeclareMathOperator{\EX}{\mathbb{E}}
\newcommand\norm[1]{\left\lVert#1\right\rVert} 

\let\oldtwocolumn\twocolumn
\renewcommand\twocolumn[1][]{%
    \oldtwocolumn[{#1}{
    \begin{center}
        \vspace{-6mm}
        \includegraphics[scale=0.3]{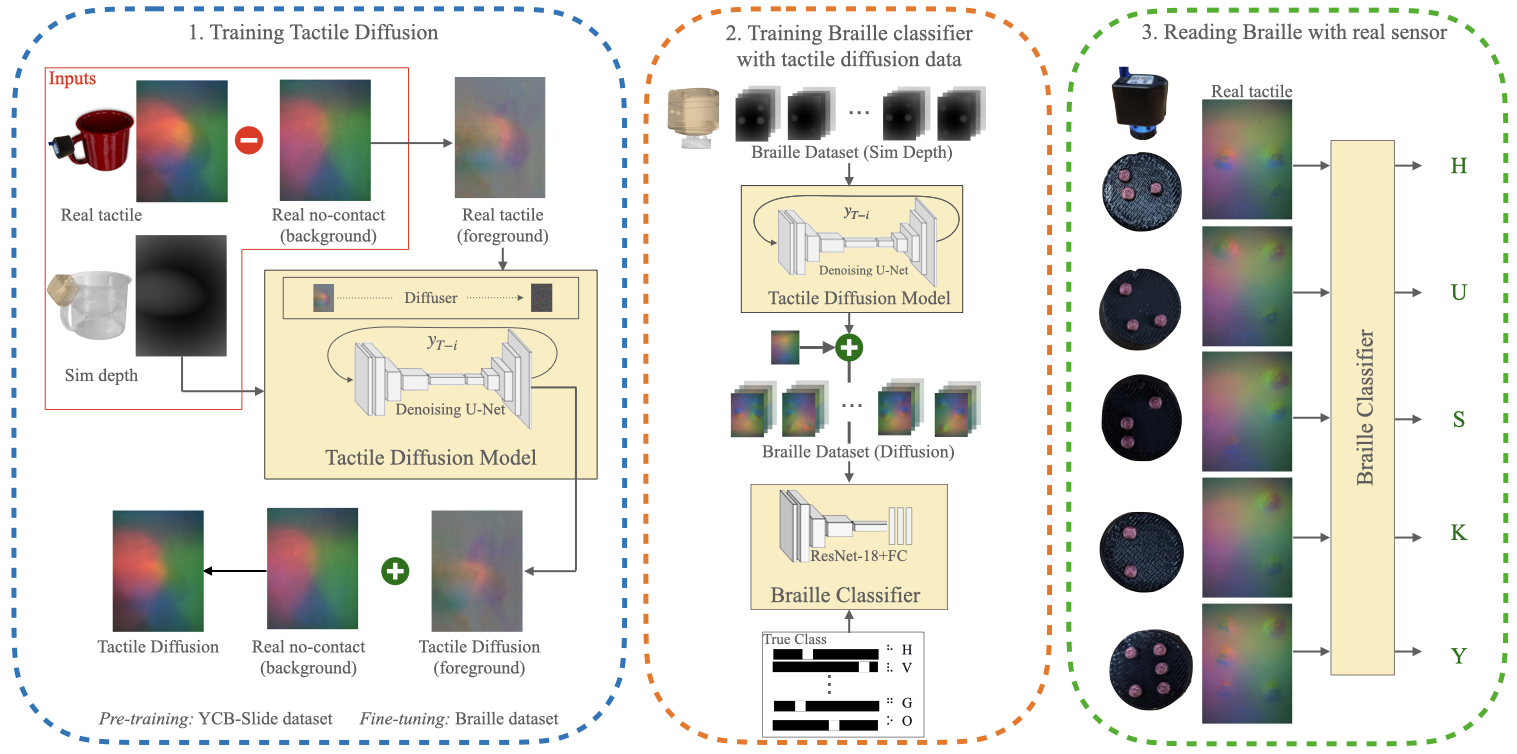}
        \vspace{-1.5mm}
        \captionof{figure}{
        (Left) Our tactile diffusion model, trained on YCB-Slide and fine-tuned on 20\% real braille, learns to generate realistic tactile images from simulated contact depth. (Middle) We train a braille classifier with data generated from tactile diffusion. (Right) On reading real braille with a DIGIT sensor this classifier outperforms classifiers trained with simulation and other approaches.
        }
        \label{fig:pipeline_training_inference}
        \vspace{-2mm}
    \end{center}
    }]
}

\title{\LARGE \bf
Learning to Read Braille:\\Bridging the Tactile Reality Gap with Diffusion Models
}

\author{Carolina Higuera$^{1}$, Byron Boots$^{1}$, and Mustafa Mukadam$^{2}$\\[2mm]
$^{1}$University of Washington, $^{2}$Meta AI
}

\begin{document}

\maketitle
\thispagestyle{empty}
\pagestyle{empty}

\begin{abstract}
Simulating vision-based tactile sensors enables learning models for contact-rich tasks when collecting real world data at scale can be prohibitive. However, modeling the optical response of the gel deformation as well as incorporating the dynamics of the contact makes sim2real challenging. Prior works have explored data augmentation, fine-tuning, or learning generative models to reduce the sim2real gap. In this work, we present the first method to leverage probabilistic diffusion models for capturing complex illumination changes from gel deformations. Our tactile diffusion model is able to generate realistic tactile images from simulated contact depth bridging the reality gap for vision-based tactile sensing. On real braille reading task with a DIGIT sensor, a classifier trained with our diffusion model achieves 75.74\% accuracy outperforming classifiers trained with simulation and other approaches. Project page: \url{https://github.com/carolinahiguera/Tactile-Diffusion}
\end{abstract}

\section{Introduction}

We present a tactile diffusion model to generate synthetic images for vision-based tactile sensors that realistically capture the illumination changes due to the elastomer's deformation. Better rendering techniques for tactile sensors will enable better feedback for a diverse range of applications such as robot manipulation, robot teleoperation, haptics, and prosthetics. A realistic rendering of tactile data can allows us to train models in simulation and deploy them in a real setting that provides feedback to the user with a virtual touch, for example, the roughness and texture features of braille characters.

Simulation of vision-based tactile sensors in conjunction with physics engines enables the possibility to quickly generate large amounts of data to prototype and debug manipulation policies and tactile models. Simulators like TACTO~\cite{wang2022tacto} and Taxim~\cite{si2022taxim} bring us closer to rendering tactile imprints and heightmaps of object-gel interactions. However, the transfer to the real system is not straightforward. In general, real tactile images are out-of-distribution in comparison with the ones in simulation. Modeling the optical response of the gel deformations and incorporating the dynamics of the contact make simulating vision-based tactile sensor very challenging. To overcome this difficulty, we can opt to rely directly on real data ~\cite{yuan2018active, dong2021tactile, she2021cable, ma2021extrinsic}, or  use data augmentation and fine-tuning ~\cite{villalonga2021tactile, gomes2021generation}.

Learning-based generative models to realistically render the in-contact indentation are also being actively explored. The problem is formulated as an image-to-image translation for which we can leverage successful solutions from the computer vision literature. The use of these learning-based models comes with several advantages, such as no need for calibration of the tactile sensor, less domain-specific knowledge about the mechanics of gel deformation, and the potential to be continuously improved. The dominant technique in the literature is training Generative Adversarial Networks (GAN) to narrow the sim2real gap. For example, cycleGAN to train sim2real and real2sim generators~\cite{chen2022bidirectional}, conditional GAN on simulated data and background~\cite{zhong2022touching}. Other works propose to do the translation in the opposite way, from real to sim using GANs, allowing  policies learned in simulation to be transferred to the real setting ~\cite{church2022tactile, lin2022tactile}.

In this work, we propose to leverage diffusion probabilistic models~\cite{ho2020denoising} to render realistic tactile images from simulation data. Based on its impressive performance in image translation ~\cite{saharia2022palette, choi2021ilvr, cheng2023adaptively, wang2022pretraining}, we leverage  these models to capture the complex illumination changes due the gel deformations in order to close the sim2real gap for vision-based tactile sensing. Diffusion models have been used in robotics to generate images of goal states for object manipulation~\cite{kapelyukh2022dall} and planning~\cite{janner2022planning, urain2022se, liang2023adaptdiffuser}, but to the best of our knowledge, they have not been explored for domain adaptation in tactile sensing. We condition the diffusion model on the simulated height map of the object-gel contact and run the denoising process to generate a tactile image that realistically renders the color gradients for the gel's deformation. 

We train our tactile diffusion model as shown in \cref{fig:pipeline_training_inference} on the \texttt{YCB-Slide} dataset~\cite{suresh2022midastouch}, which gives us access to a diverse set of contact interactions between DIGIT sensor and YCB objects. Then, we fine-tune our model with a few samples of real braille data, our downstream task domain.  We show how our tactile diffusion model facilitates (i) the rendering of realistic tactile simulation images that highlights the object's texture features,  which allows (ii) achieving a zero-shot transfer on real braille reading task with a DIGIT sensor. The braille classifier trained with our diffusion model achieves 75.74\% accuracy outperforming classifiers trained with simulation and other approaches.


\section{Related Work}

\subsection{Vision-based tactile simulation}
Simulation of vision-based tactile sensors can speed up learning policies for contact-rich manipulation tasks in robotics. This is currently an active area of research facing two main challenges. First, how to capture the dynamics of the contact and second, how to model the optical and illumination properties of the sensor under deformation conditions, to obtain realistic perceptual images. TACTO~\cite{wang2022tacto} aims to tackle specifically the second challenge, while leaving the simulation of contact forces to the underlying physics engine, such as PyBullet~\cite{coumans2016pybullet}. This simulator  synchronizes the poses of objects and sensors in the rendering engine and fetches the rendered tactile and depth imprints. Results of experiments for the DIGIT~\cite{lambeta2020digit} and OmniTact~\cite{padmanabha2020omnitact} sensors suggest the need for techniques such as data augmentation or domain adaptation to narrow the sim2real gap. 

Taxim~\cite{si2022taxim} simulates  with a second-order polynomial look-up table the optical response of a Gelsight sensor~\cite{yuan2017gelsight}. Calibration against real data is needed to fit the parameters, a procedure that has to be done per sensor. A height map is constructed from the object's shape in the contact area, smoothing the edges with pyramid Gaussian kernels. The normal value is extracted for each point in the height map and mapped to an intensity value through the look-up table to render the tactile image. \cite{si2022grasp} explores how to use Pybullet and Taxim to capture both contact forces and tactile data for a grasp stability task.

\cite{xuefficient} proposes a different approach, focusing on the challenge of capturing the dynamics of the contact between object and sensor. Instead of rendering tactile images, they simulate both the normal and shear forces covering the contact surface of the sensor. The method consists of representing the sensor pad as a set of points and using a penalty-based tactile model to characterize the force on each one. Experiments show that they can approximate the elastomer of a GelSlim sensor~\cite{donlon2018gelslim} generating dense tactile force fields.

Some attempts into narrowing the sim2real gap use physics-based rendering~\cite{agarwal2021simulation} or numerical methods to model the deformation of the elastomer. For example, ~\cite{gomes2021generation}  uses Bivariate 2D Gaussian filtering  followed by the Phong shading model~\cite{phong1975illumination} for rendering the sensor's internal illumination.  The Structural Similarity (SSIM) reported was around 0.859 for a GelSight sensor; however, parameters for the reflectance function and for the Gaussian kernel have to be set manually.

\subsection{Vision-based tactile synthetic image generation}

Data-driven approaches to simulate realistic tactile images require less domain-specific knowledge and can keep being improved in an online fashion. To overcome the challenges of simulating the color gradients that characterize the deformation of the membrane due to object indentation, Generative Adversarial Networks (GAN)~\cite{goodfellow2020generative} have been popular among the tactile sensing community as a domain adaptation technique. The main idea is to use the network as a translator to make the simulation images resemble the style of the real data.~\cite{chen2022bidirectional} proposes to use a CycleGAN trained with unpair sim and real data to obtain sim2real and real2sim generators.~\cite{zhong2022touching} trains a conditional GAN to generate tactile data conditioned on both the RGB-D images rendered by a NeRF model and a reference background image for the tactile sensor. The SSIM reported was 0.865 with respect to ground truth images from a DIGIT sensor. Approaches for real2sim are being explored as well. For example, ~\cite{church2022tactile, lin2022tactile} translate real to simulation images and use them as states in an RL policy for tasks such as  edge and surface following, object rolling, and pushing. They evaluate their approach for three sensors TacTip, DIGIT, and DigiTac, reporting an SSIM above 0.98 for the edge and surface trajectories.

\subsection{Diffusion models in robotics}
Given the recent success of diffusion probabilistic models in computer vision~\cite{ho2020denoising, ramesh2022hierarchical, nichol2021glide, Kim_2022_CVPR, rombach2022high, saharia2022photorealistic}, researchers have begun exploring their advantages in robotics. \cite{kapelyukh2022dall} proposes their usage in manipulation tasks to generate goal-state images of a human-like arrangement of objects from a text description.~\cite{liu2022structdiffusion} uses diffusion models to complex multi-step 3D planning tasks for object rearrangement.~\cite{janner2022planning} introduces a trajectory-level diffusion probabilistic model, a  method capable of planning trajectories for several tasks, by using different guidance functions.~\cite{urain2022se} focuses on learning SE(3) diffusion models for 6DoF grasping to represent grasp pose distributions as cost functions.~\cite{liang2023adaptdiffuser} introduces an evolutionary planning method with diffusion for offline reinforcement learning, that can self-improve by using guidance from reward gradients. To the best of our knowledge this work is the first to employ diffusion models for tactile sensing.

\section{Tactile Diffusion Model}
In this paper, we focus on the problem of closing the sim-to-real gap rendering images for vision-based tactile sensors. Simulators such as TACTO~\cite{wang2022tacto} and Taxim~\cite{si2022taxim}, among others, can help to accelerate the learning of models to extract useful information from contacts and learning policies for contact-rich manipulation tasks. However, the sim2real transfer of those models can be challenging due to optical differences between real and simulated images, given that these simulators cannot completely resemble the complexity and non-ideal light transmission on the elastomer of gelsight-like sensors. Our goal is to use diffusion models to generate realistic synthetic images for the DIGIT sensor~\cite{lambeta2020digit} conditioned on depth images from simulation. We explore the capabilities of diffusion models to capture the internal illumination and the membrane deformations of the sensor caused by the indentation of the object in contact with the elastomer. 

\subsection{Training tactile diffusion models}
We view the problem of generating realistic synthetic tactile images as image-to-image translation. Given aligned pairs of sim and real tactile images, our goal is to transform an input grayscale depth image from a tactile simulator to a plausible color image that resembles the color gradients around the in-contact indentation. For doing so, we follow the approach proposed in~\cite{saharia2022palette} for image colorization.

Conditional diffusion models allow us to formulate the denoising process as $p(y|x)$, where $x$ is the image that we are using for conditioning or querying the model and $y$ is the resulting color image. Our goal is to learn the reversed process to recover the ground-truth tactile image $y_0$ which contains information about the in-contact indentations, conditioning the model on the corresponding depth image $x$ from a tactile simulator. 

During the forward process, the original image $y_0$ is diffused by adding noise $\epsilon_t$ during $t \in T$ timesteps. This noise is drawn from a standard normal distribution and the amount of noise that is blended into the image is controlled by the hyperparameter $\beta_t$:
\begin{equation}
    \begin{split}
        \hat{y}_1 &= \sqrt{1-\beta_1}y_0 + \sqrt{\beta_1}\epsilon_1 \\
        \hat{y}_t &= \sqrt{1-\beta_t}\hat{y}_{t-1} + \sqrt{\beta_t}\epsilon_t \;\;\; \forall t \in  2 \cdots T 
    \end{split}
\end{equation}

However, it is possible to directly draw samples $\hat{y}_t$ by iterative substituting the above equations, as:
\begin{equation} \label{eq_yt}
    \hat{y}_t = \sqrt{\alpha_t}y_0 + \sqrt{1-\alpha_t}\epsilon, \;\;\; \text{where  } \alpha_t=\prod_{s=1}^{t}1-\beta_s
\end{equation}

The decoder does the reverse process, mapping back from $\hat{y}_t$ to $\hat{y}_{t-1}$ until recovering the original image $y_0$, while conditioned on $x$. For doing so, a decoder $f_\theta(x,\hat{y}_t,\alpha_t)$ with parameters $\theta$ is trained to recover the noise vector $\epsilon_t$ added to the image $\hat{y}_t$. For training the decoder, we optimize an objective resembling denoising score matching \cite{ho2020denoising}. Our loss function corresponds to the following objective:
\begin{equation}
    L(\theta) = \EX_{(x,y_0)}\EX_{\epsilon, t}\left[ \norm{\epsilon - f_\theta(x,\hat{y}_t,\alpha_t)}^2 \right]
\end{equation}

\subsection{Inference from sim-depth as conditional image}
The forward process of the encoder is constructed so conditional distribution $p(\hat{y}_T|y_0, x)$ and marginal distribution $p(\hat{y}_T|x)$ of the  final latent variable $\hat{y}_T$ both become the standard normal distribution. Therefore, we can start the inference process with pure Gaussian noise, followed by $T$ iterative steps to map back to the gel deformation $y_0$.

After reparameterizing the network to predict the added noise, we can approximate the initial image $y_0$ by rearranging terms in~\cref{eq_yt}:
\begin{equation}\label{eq_approx_y0}
\begin{split}
    y_0 &\approx \frac{1}{\sqrt{\alpha_T}} \left[ y_T - \sqrt{1-\alpha_T}\hat{\epsilon} \right]\\
    &=  \frac{1}{\sqrt{\alpha_T}} \left[ y_T - \sqrt{1-\alpha_T}f_\theta(x,\hat{y}_T,\gamma) \right]
\end{split}
\end{equation}


By Bayes's rule, we can get the conditional diffusion distribution $q(\hat{y}_{t-1}| \hat{y}_{t}, y_0)$ to recover the image prior to adding noise, which corresponds to the normal distribution:
\begin{equation} \label{eq_diff_dist}
\small
    \mathcal{N}\left[ \frac{(1-\alpha_{t-1})}{1-\alpha_t}\sqrt{1-\beta_t}\hat{y}_t + \frac{\sqrt{\alpha_{t-1}}\beta_t}{1-\alpha_t}y_0, \frac{\beta_t(1-\alpha_{t-1})}{1-\alpha_t}\mathbf{I} \right]
\end{equation}

After plugging in the approximation of $y_0$ from~\cref{eq_approx_y0} into the conditional diffusion distribution in~\cref{eq_diff_dist}, each iteration of the reverse process can be recovered as:
\begin{equation}
    \hat{y}_{t-1} = \frac{1}{\sqrt{1-\beta_t}}\hat{y}_t - \frac{\beta_t}{\sqrt{1-\alpha_t}\sqrt{1-\beta_t}}\hat{\epsilon} + \sigma\epsilon
\end{equation}
where $\sigma\epsilon$ accounts for the variance of the conditional diffusion distribution and it is drawn from a standard normal distribution \cite{prince2023understanding}.


\begin{figure*}[t]
    \centering
    \includegraphics[scale=0.24]{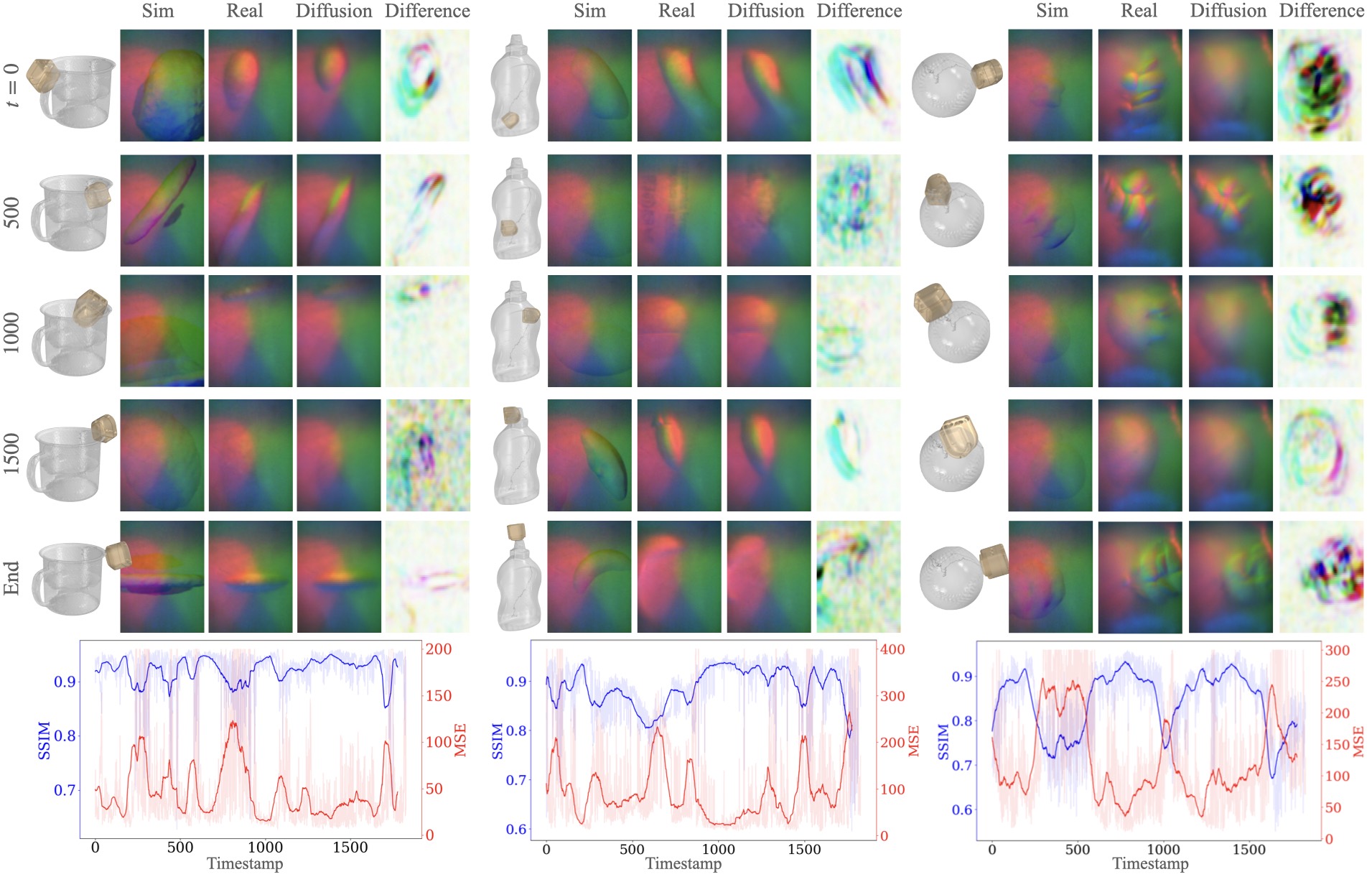}
    \vspace{-3mm}
    \caption{Simulation, real, tactile diffusion, and difference with respect to real images at different timestamps in a trajectory of contact interactions between DIGIT sensor and YCB objects (left: mug, middle: mustard\_bottle, right: baseball) from the YCB-Slide dataset. Images generated with tactile diffusion have a similar structure as the real ones (SSIM is generally above 0.80). The differences are mostly due to small misalignment of the contact patch or missing rendering specific texture information that is captured by the sensor's camera, such as the label's text on the mustard\_bottle.}
    \label{fig:ycb_diffusion}
    \vspace{-5mm}
\end{figure*}

\section{Experiments}

In this section, we evaluate tactile diffusion for braille reading task, as shown in~\cref{fig:pipeline_training_inference}. First, we pre-train the tactile diffusion model on a rich and diverse dataset of sliding interactions between a vision-based tactile sensor and standard YCB objects from the YCB-Slide dataset~\cite{suresh2022midastouch}. Then, we fine-tune our model with a few real samples of braille tactile images. We use tactile diffusion to generate realistic tactile images from simulation data to train a braille classifier. We consider reading braille an interesting and natural tactile task that elucidates the importance of rendering realistic object textures in tactile simulation for building tactile models for downstream tasks. We demonstrate that tactile diffusion allows us to achieve a zero-shot transfer of our classifier to read real braille text with a DIGIT sensor.


\subsection{Tactile diffusion pre-training}

We use the \texttt{YCB-Slide} dataset~\cite{suresh2022midastouch}, which consists of sliding contact interactions between the DIGIT sensor and 10 YCB objects. The dataset contains real tactile images, sensor poses, and ground-truth mesh models for 5 trajectories per object with a duration of 60s. With the sensor poses and the object meshes, we use the TACTO simulator to recreate each trajectory and collect the simulated depth image for each timestep. Our dataset for training the diffusion model consists of 180k aligned pairs of sim depth and real tactile images, split into $80\%$ for training and $20\%$ for testing. 

The decoder of the tactile diffusion model uses a conditional U-Net backbone with  2 ResNet blocks for each stage of down-sampling and up-sampling respectively. We allow conditioning on the sim depth image via concatenation, following~\cite{saharia2022palette, saharia2022image}. We use a linear noise scheduler of $(1e^{-4}, 0.02)$ with $T=500$ timesteps for training and inference. We use Adam~\cite{kingma2014adam} optimizer with a learning rate of $1e^{-4}$.

For training our tactile diffusion model, we follow the pipeline shown in~\cref{fig:pipeline_training_inference} (left). We pre-process the ground-truth data by doing a foreground extraction. This consists of subtracting from all images a real no-contact image from the sensor. Different DIGIT sensors might have a slight variation on the background due to manufacturing, so this makes the diffusion model agnostic to the background and focuses only on the color gradients that happen around the gel deformations. An illustration of the pipeline for training and inference of the diffusion model is shown in~\cref{fig:pipeline_training_inference}. For inference, we start the denoising process from pure Gaussian noise with the sim depth image as input to condition the model. The intuition is to query the model about how the sim depth image will look when using a real sensor. After inference, we post-process the generated foreground by adding the background back. We can add as background no-contact image from any DIGIT sensor that we will use in a downstream task. For training and inference we use a RTX-3080 GPU and the inference time of a batch of 30 images is 23 seconds.

To evaluate the performance of the tactile diffusion model, we compute the SSIM (structural similarity) and MSE (mean squared error) pixel-wise on test data. This dataset corresponds to trajectories from the \texttt{YCB-Slide} dataset that were not used for training the diffusion model. In~\cref{fig:ycb_diffusion} we show some snapshots of sim, real and generated tactile images for a DIGIT sensor, as well as their differences. For the case of the mug and mustard\_bottle, the SSIM for the generated images is above 0.80 across the trajectory. This means that the diffusion model is capable of generating tactile images with similar structure and gel deformation luminance information in comparison to images from a real sensor. The differences for the case of the mug are mostly small misalignment of the contact patch; for the mustard\_bottle, in addition to misalignment, the model does not accurately render very specific texture information that is captured by the sensor's camera (such as the label's text in $t=500$). However, this kind of texture does not provide any useful information about the contact, even though it negatively impacts the metrics, especially the MSE, in terms of the similarity of images. The trajectory for the baseball is more difficult for the diffusion model to render with fine detail from simulation data. When comparing the real and generated tactile image we can notice that the real sensor captures a lot of texture information that the diffusion model generalizes as a contact path. Although in terms of SSIM and MSE the images have a lower similarity, the diffusion model captures the contact patch accurately.

\subsection{Tactile diffusion fine-tuning}

We fine-tune our tactile diffusion model with pairs of (sim depth, real) images from braille contact interactions. Our datasets for fine-tuning and testing consist of real tactile data collected from a DIGIT sensor when in contact with 27 3D-printed braille characters (letters A-Z and \#). Each object has size $2\times2\times1$cm. The setup for collecting the data consist of a Franka Panda arm with a DIGIT sensor mounted on a custom 3D-printed flange so the sensor faces perpendicular to the end-effector. Taking as reference the center of the object, the contacts are located on a 3D grid of $3\times3\times4$ with steps of $\pm 5$mm on the XY axis and $-2$mm on the Z axis for yaw rotations of $0 ^{\circ}, \pm45 ^{\circ} $ and $\pm90 ^{\circ} $. This results in 180 images per object with a resolution of $320\times240$. Additionally, we replicate the same data collection using PyBullet and TACTO to obtain the depth penetration images for each interaction sensor's gel and object. Our dataset consists of 1080 pairs of fine-tuning images and 837 pairs of test images.

\begin{figure}[t]
    \centering
    \includegraphics[scale=0.35]{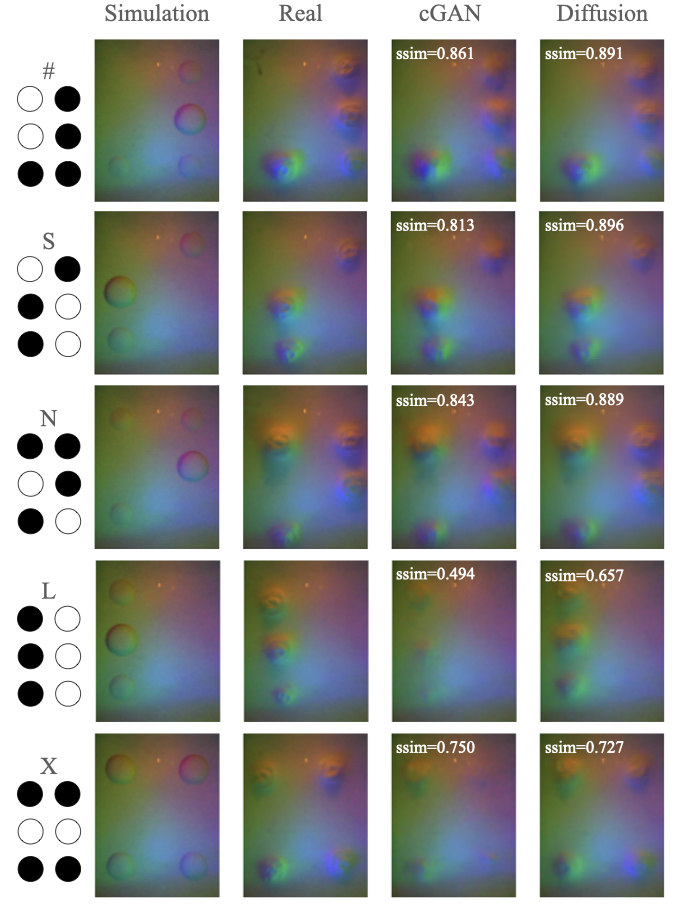}
    \vspace{-2mm}
    \caption{Example comparisons of sim, real, cGAN, and tactile diffusion images from the braille test dataset.
    Tactile diffusion consistently generates images with higher SSIM with respect to the real sensor image. Notice that tactile diffusion does not skip the representation of any dots in the braille character.}
    \label{fig:braille_test}
    \vspace{-5mm}
\end{figure}

We fine-tune the tactile diffusion model previously trained on the \texttt{YCB-Slide} with $20\%$ of the fine-tuning dataset. In~\cref{fig:braille_test} we show, for different letters, samples of sim, real, and generated tactile images by tactile diffusion and cGAN as a baseline. Although the use of cGAN to generate tactile images is common in the literature, the models are not open-sourced, thus we we are using our own implementation of cGAN, conditioning  on the same sim depth image.

Qualitatively, tactile diffusion is able to represent the corresponding gel deformation by rendering the color changes around the object indentations with high detail level. These indentations correspond to the bumps that characterize each braille character. cGAN is able to do it as well for some characters, but exhibits errors rendering the level of gel deformation or skipping its representation. This is congruent with the findings in~\cite{dhariwal2021diffusion}, where the authors explain that GANs are able to trade off diversity for fidelity, producing high-quality samples, but not covering the whole distribution. In general, we found cGAN more difficult to condition and exhibit less level of texture detail in comparison with the images generated by the diffusion model. Our tactile diffusion model also presents some differences with respect to the real image. However, these differences mostly correspond to misalignment in the location of the bump's indentation.

To assess quantitatively both models, we compute the SSIM and MSE pixel-wise between real and generated images on our test dataset of real tactile images. Our tactile diffusion model has an SSIM of $0.908$ and MSE of $36.02$  pixel-wise (maximum pixel value is $255$ per channel); whereas cGAN has an SSIM of $0.879$ and MSE of $47.99$. These metrics corroborate that the tactile diffusion model can generate better synthetic images than cGAN and that these images are structurally similar to those collected with a real DIGIT sensor for this braille dataset.

\subsection{Tactile diffusion for reading braille}

To evaluate if processing the simulation data with the tactile diffusion model facilitates the sim2real transfer, we train a classifier to detect the corresponding letter for reading braille text with a vision-based tactile sensor.  We collect a separate dataset using TACTO of depth images from contact interactions of the DIGIT sensor with each of the meshes of the 27 braille characters. Unlike the data used for fine-tuning, in this dataset the sensor is always placed in a vertical position to read Braille, however, we take into account small variations in the orientation of the sensor. The contacts are located on a 3D grid of $3\times3\times4$ with steps of $\pm 3$mm on the XY axis and $-2$mm on the Z axis for yaw rotations from $-25^{\circ}$ to $+25^{\circ}$ with steps of $5^{\circ}$. 

We trained this classifier using all current common approaches for tackling sim2real when using vision-based tactile sensing. This consists of training the model using data from sim + data augmentation (lighting randomization), sim + fine-tuning with real data, sim + data augmentation + fine-tuning  with real data, cGAN, and tactile diffusion. For the last two approaches, we performed data adaptation on the sim depth images, following the pipeline in~\cref{fig:pipeline_training_inference} (middle). \cref{tbl:accuracy_clf} shows the performance of all these models (accuracy, precision, and recall) when testing the models on our test braille dataset of 837 images.

\begin{table}[!t]
\caption{Metrics on braille classification task.}
\vspace{-1mm}
\label{tbl:accuracy_clf}
\begin{tabular}{lcccc}
\toprule
\begin{tabular}[l]{@{}l@{}} \textbf{Training data} \\ \textbf{source} \end{tabular} &
\begin{tabular}[c]{@{}c@{}} \textbf{\% real data} \\ \textbf{fine-tuning} \end{tabular} &
\begin{tabular}[c]{@{}c@{}} \textbf{Accuracy} \\ \textbf{\%} \end{tabular} &
\textbf{Precision} &
\textbf{Recall} \\
\midrule
\multirow{4}{*}{Sim}                                                    & -   & 30.23 & 0.34 & 0.30 \\
                                                                        & 20  & 64.99  & 0.71 & 0.65 \\
                                                                        & 80  & 73.11 & 0.80 & 0.73 \\
                                                                        & 100 & 73.95 & \textbf{0.81} & 0.74 \\
\hdashline
\multirow{2}{*}{Sim + data aug.}                                        & -   & 43.48 & 0.61 & 0.43 \\
                                                                        & 100 & 73.23 & 0.76 & 0.73 \\
\hdashline
cGAN & -   & 31.18 & 0.40 & 0.31 \\
Tactile diffusion & - & \textbf{75.74} & 0.79 & \textbf{0.76} \\
Real                                                                    & -   & 100.0 & 1.00 & 1.00 \\
\bottomrule
\multicolumn{5}{c}{\scriptsize Training cGAN on 100\% real, tactile diffusion on YCB-Slide + 20\% real}
\end{tabular}
\vspace{-6mm}
\end{table}


Using real data to train the downstream task would be the ideal as we guarantee that both train and test data are under the same distribution. The braille classifier trained directly on real data can perfectly distinguish the letters based on the imprints of the braille characters on the sensor's gel. However, collecting real data is expensive and time-consuming and prohibitive to scale. In practice, we would like to be able to train the downstream task solely in simulation and achieve good performance when deployed in real. Transferring directly the model trained on raw simulation highlights the sim2real gap when working with vision-based tactile data. Data augmentation improves the generalization of the model but not enough to induce the distribution of the real data on the training dataset. Fine-tuning the model on real data definitely helps to improve its performance. We investigate when most performance can be achieved with the least amount of real fine-tuning data. Tactile diffusion is trained on a general dataset of contacts from YCB-Slide and fine-tuned with only $20\%$ of the task relevant braille data. Under these conditions, it  achieves a zero-shot accuracy of $75.74\%$ on real tactile data. Fine-tuning the simulation model to achieve similar performance requires collecting 4 times more real tactile data.

Comparing data adaptation techniques, tactile diffusion has significantly better performance than cGAN.  This is expected since cGAN sometimes skips the representation of bumps for some braille characters. These anomalies in the tactile image can instead represent a different character leading to missclassification, which hurts the downstream task.  As shown in other problems of image synthesis~\cite{dhariwal2021diffusion}, going from pure noise to a realistic image in one step is a possible cause of the noisy behavior of cGAN. Overall, these results highlight the promising future of tactile diffusion to close the sim2real gap for vision-based tactile sensors.
 
\begin{figure}[t]
    \centering
    \includegraphics[scale=0.26]{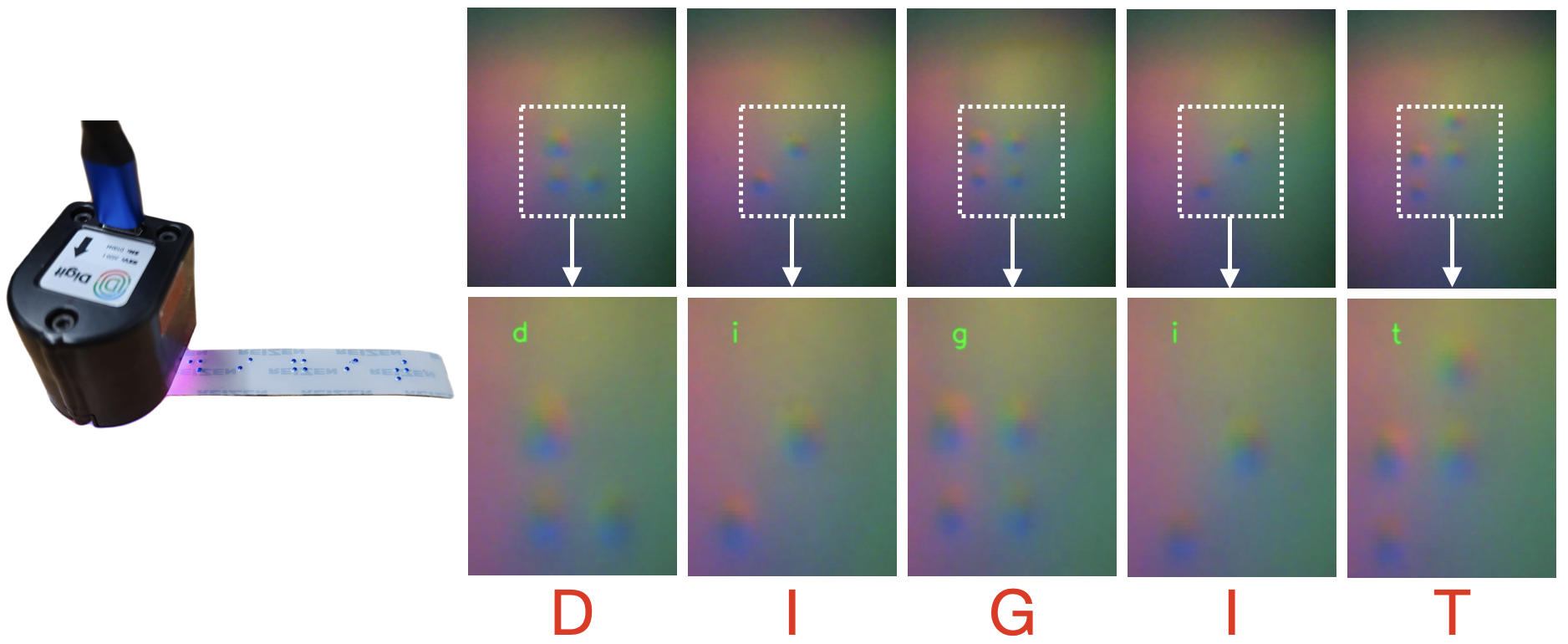}
    \vspace{-3mm}
    \caption{We type the word \textit{DIGIT} on a vinyl tape with a braille label maker. The tape is read with a DIGIT sensor using the classifier trained with tactile diffusion. This classifier generalizes to slight variations of the braille imprints for real-sized characters.}
    \label{fig:reading_braille}
    \vspace{-6mm}
\end{figure}

We tested the classifier trained on tactile diffusion data for reading real-sized braille text with the DIGIT sensor. We type in a vinyl tape the word \textit{DIGIT} with a braille label maker. We do a zoom-in of the braille indentation on the gel, in order to approximate the size of the whole braille pattern that the classifier was trained on. In~\cref{fig:reading_braille} we show snapshots of the classifier trained with tactile diffusion zero-shot reading real-sized braille.

\section{Discussion}

\textbf{Summary.} In this work, we present the application of diffusion probabilistic models to generate realistic tactile images for vision-based tactile sensors, conditioned on simulation data. Our model takes as input a depth image from simulation that represents the height map of the gel deformation caused by the object's indentation. Conditioned on this data, the model generates a foreground image that realistically renders the color gradients for the gel's deformation. Then, as a post-processing step, we can add the no-contact background image from a real sensor. This allows the method to be robust against slight variations between sensors of the same type as a consequence of the manufacturing variability of the elastomer. Our experiments demonstrate that the generated tactile images resemble a similar structure in comparison to real ones when simulating a trajectory of a DIGIT sensor on the surface of YCB objects. Moreover, we show that the tactile diffusion model allows training models solely on simulation data and their posterior transfer to real scenarios in downstream tasks like braille classification. We believe that diffusion models are a promising tool to close the sim2real gap for vision-based tactile sensing.

\textbf{Limitations.} The tactile diffusion model can be used to post-process a dataset of simulation data to train a model for a downstream task in an offline fashion similar to our braille classifier. The main limitation of the current integration of the diffusion model in the simulation loop is the compute time to carry out the denoising process. To speed up the inference, alternatives can be explored like running the diffusion process in latent space rather than on pixel space~\cite{rombach2022high} and denoising diffusion with GANs~\cite{xiao2022DDGAN}.

\section*{Acknowledgment}
The authors thank Sudharshan Suresh, Mike Lambeta, and Roberto Calandra for help with TACTO simulator and DIGIT sensors.

\balance
\bibliographystyle{IEEEtran}
\bibliography{IEEEabrv, references}

\end{document}